\documentclass[sigconf]{acmart}

\usepackage{times}
\usepackage{graphicx} 
\usepackage{subfigure} 

\usepackage{natbib}

\usepackage{algorithm}
\usepackage[noend]{algorithmic}

\usepackage{hyperref}
\usepackage[utf8]{inputenc} 
\usepackage[T1]{fontenc}    
\usepackage{hyperref}       
\usepackage{url}            
\usepackage{booktabs}       
\usepackage{amsfonts}       
\usepackage{nicefrac}       
\usepackage{microtype}      
\usepackage{bm}
\usepackage{bbm}
\usepackage{mathtools}
\usepackage{amsmath}
\usepackage{color} 
\usepackage{dsfont}
\usepackage{booktabs} 
\usepackage{multirow}
\usepackage{dblfloatfix} 

\DeclareMathOperator*{\argmax}{argmax}
\setcopyright{rightsretained}

\acmConference{KDD'17}{August 13 - 17, 2017}{Halifax, Nova Scotia - Canada} 
\acmYear{2017}
\copyrightyear{2017}

\acmPrice{15.00}
\begin{document}
\title{
Regularizing Model Complexity and Label Structure for Multi-Label Text Classification}

\author{Bingyu Wang}
\affiliation{%
  \institution{Northeastern University}
  \city{Boston} 
  \state{MA} 
  \postcode{02115}
}
\email{rainicy@ccs.neu.edu}

\author{Cheng Li}
\affiliation{%
  \institution{Northeastern University}
  \city{Boston} 
  \state{MA} 
  \postcode{02115}
}
\email{chengli@ccs.neu.edu}

\author{Virgil Pavlu}
\affiliation{%
  \institution{Northeastern University}
  \city{Boston} 
  \state{MA} 
  \postcode{02115}
}
\email{vip@ccs.neu.edu}

\author{Javed Aslam}
\affiliation{%
  \institution{Northeastern University}
  \city{Boston} 
  \state{MA} 
  \postcode{02115}
}
\email{jaa@ccs.neu.edu}


\begin{abstract}


Multi-label text classification is a popular machine learning task where each document is assigned with multiple relevant labels. This task is challenging due to  high dimensional features and correlated labels. Multi-label text classifiers need to be carefully regularized to prevent the severe over-fitting in the high dimensional space, and also need to take into account label dependencies in order to make accurate predictions under uncertainty. Most of the existing multi-label classifier designs focus on incorporating label dependencies into model training phase and optimize a strict set-accuracy measure; in practice a partial reward like F-measure makes more sense for set labels such as tags, ontologies, medical codes, movie genres, etc. 

We demonstrate significant and practical improvement by carefully regularizing the model complexity during training phase, and also regularizing the label search space during prediction phase. Specifically, we regularize the classifier training using \emph{Elastic-net} (L1+L2) penalty for reducing model complexity/size, and employ \emph{early stopping} to prevent overfitting. At prediction time, we apply \emph{support inference} to restrict the search space to label sets encountered in the training set, and F-optimizer \emph{GFM} to make optimal predictions for the F1 metric. We show that although support inference only provides density estimations on existing label combinations, when combined with GFM predictor, the algorithm can output unseen label combinations. 

Taken collectively, our experiments show state of the art results on many benchmark datasets. Beyond performance and practical contributions, we make some interesting observations. Contrary to the prior belief, which deems support inference as purely an approximate inference procedure, we show that support inference acts as a strong regularizer on the label prediction structure. It allows the classifier to take into account label dependencies during prediction even if the classifiers had not modeled any label dependencies during training.

\end{abstract}

%
%




\maketitle

%

\section{Introduction}

Classifying a document into one of the predefined categories has been well studied, and many multi-class classifiers can be applied to solve this problem. In practice, however, text documents are often naturally associated with more than one label, and it is desirable to find all the relevant labels for the document. For example, in a medical note, a patient may present multiple illnesses or undergo a procedure with multiple billing codes; in news categorization, an article can be associated with multiple topics. This problem can be formulated as a multi-label classification task, which seeks to predict a \emph{subset} of possible labels for a data object.

Formally, in a multi-label classification problem, we are given a set
of label candidates $\mathcal{L}=\{1,2,...,L\}$. Every data point
$\mathbf{x}\in \mathbb{R}^D$ matches a subset of labels
$\mathbf{y}\subseteq \mathcal{L}$, which is often also written in the
form of a binary label vector $\mathbf{y}\in \{0,1\}^L$, with each bit $y_l$
indicating the presence or absence of the corresponding label.
The goal of learning is to build a classier $h: \mathbb{R}^D\mapsto
\{0,1\}^L$ which maps an instance to a subset of labels. The label
subset $\mathbf{y}$ can be of arbitrary size (written as  $|\mathbf{y}| = ||\mathbf{y}||_1$). Multi-label generalizes binary and multi-class: when the size is
restricted to be~1, the problem is called multi-class; if the total number of label
candidates $L$ is~2, the problem is binary classification.


\subsection{Challenges in Multi-label Text Classification}
Multi-label text classification is a challenging task for at least two reasons: 1) the labels exhibit complex dependency structures, 2) the features are high dimensional and sparse.
Text labels  labels such as \texttt{election} and \texttt{politics} are \emph{dependent} variables in general, so an
independent prediction (also called Binary Relevance) is unlikely to work well \cite{tsoumakas2006multi}. Labels can be
\emph{many}, so learning approaches dealing explicitly with each of
the exponential number of label subsets (e.g. PowerSet method \cite{tsoumakas2006multi} ) are infeasible; even when
feasible, they suffer from scarce data and limited label subsets
observed during training. In recent years, there have been an growing interest in developing multi-label methods that are capable of modeling label dependencies while at the same time avoid the exponential computational complexity. Examples include probabilistic classifier chains \cite{read2011classifier}, 
conditional random fields \cite{ghamrawi2005collective} , and conditional Bernoulli mixtures \cite{li2016conditional}. 


The second challenge in multi-label text classification is that the commonly used bag-of-words feature representation is high dimensional and sparse. For example, the WISE dataset has 301,561 unigram features and 203 labels. If ngram features are further included, the feature size would grow dramatically (by at least a factor of 10). Model training without careful regularization can lead to severe over-fitting. This is especially a problem for high capacity classifiers which aim to model complex label dependencies, such as conditional Bernoulli mixtures \cite{li2016conditional} which can get very high training accuracy without improving generalization ability. The large number of features, together with the large number of labels, also make the produced classifiers large in size, as measured by the storage space they occupy. Even for the most simplistic binary relevance method, which allocates one logistic regression for each label, the model size grows linearly with the product of the number of labels and number of features. For more sophisticated methods, the model sizes will grow even faster. The training regularization steps (elastic-net, early stopping) are addressing these issues directly; we used well established ideas for both of these; but we provide an efficient and public implementation tailored to specific multi-label algorithms, analyze the specific effects, and draw novel conclusions. 



\subsection{F1 Metric and Optimal F Predictions}
The most widely used evaluation measure for multi-label text classification is the F1 metric \cite{waegeman2014bayes}, which assigns partial credit to
``almost correct'' answers and handles label imbalance well. Let
$\mathbf{x}$ be an instance (document)
with ground truth label vector $\mathbf{y}$,  and $\mathbf{y'}$ be a
prediction label vector made. \textbf{F1-metric}  is defined as
\begin{align}
F(\mathbf{y},\mathbf{y'})=\frac{2\sum_{l=1}^L y_l y'_l}{\sum_{l=1}^Ly_l +\sum_{l=1}^Ly'_l},
\end{align}
which can also be written as the harmonic mean between precision and recall:
\begin{align}
Precision(\mathbf{y},\mathbf{y'})=\frac{\sum_{l=1}^L y_l y'_l}{\sum_{l=1}^Ly'_l} \hspace{5ex}
Recall(\mathbf{y},\mathbf{y'})=\frac{\sum_{l=1}^L y_l y'_l}{\sum_{l=1}^Ly_l}
\end{align}

The reported F1 on a dataset is the average of per-instance F1 values. There exist a few methods which explicitly take into account the F1 metric during training \cite{parambath2014optimizing, gasse2016f, pillai2017designing}, but most of the popular methods that provide a joint estimation in the form of $p(\mathbf{y}|\mathbf{x})$ are trained by standard maximum likelihood estimation without considering F1 metric as objective. For such methods, it is still possible to use an F1 optimal prediction strategy post-training, that is, output $\mathbf{y^*}$ which maximizes the expected F1:
\[
\mathbf{y^*}=\arg\max_\mathbf{y'} \mathbf{E}_{\mathbf{y}\sim p(\mathbf{y}|\mathbf{x})} [F(\mathbf{y},\mathbf{y'})]=\arg\max_\mathbf{y'} \sum_{\mathbf{y}} p(\mathbf{y}|\mathbf{x})\cdot F(\mathbf{y},\mathbf{y'})
\]

The General F-measure Maximizer(\textbf{GFM}) algorithm \cite{waegeman2014bayes} is an efficient algorithm that finds the F1 optimal prediction for a given instance based on some probability estimations. The GFM algorithm does not work directly with a joint estimation  $p(\mathbf{y}|\mathbf{x})$, but rather, some $L^2$ marginal distributions (which will be defined more precisely in Section \ref{sec:pred}). The paper~\cite{waegeman2014bayes} proposed two ways of getting these $L^2$ marginals (or probabilities per testing instance): (1) a new classifier that directly estimates these marginals from data; (2) use a probabilistic joint classifier/estimator  $p(\mathbf{y}|\mathbf{x})$ and sampling to generate the required $L^2$ probabilities.

We find that option (1) is a very difficult, if not unsolvable in practice, problem --- although it is indeed appealing as a theoretical exercise. We instead develop an efficient solution based on (2) but make a critical change: we first train a joint estimator  $p(\mathbf{y}|\mathbf{x})$, and then derive the required marginals using an approximation called \textbf{support inference}. These marginals are then fed into GFM to produce the F1 optimal prediction. We show that the proposed solution is reasonably simple to implement, and very effective. In the experimental section we analyze the support inference procedure as a strong regularizer on the label structures: the final optimal-F prediction takes into account label dependencies even if the joint estimation had not modeled any label dependencies during training, and even if the optimization objective for the joint was not related to F.

In summary, we are using several previous ideas to put together a good regularization for both features/training and labels/testing. Perhaps the biggest points we make are the new explanations of what works and why. Specifically we make the following contributions:

$\bullet$ regularize multi-label text classifier training use Elastic-net penalty (L1 + L2) and early stopping, in order to avoid over-fitting in the high dimensional space, as well as to reduce the model size.

$\bullet$ regularize multi-label prediction by combining support inference and GFM predictor. Support inference allows GFM to  work conveniently with well developed probabilistic classifiers, and GFM outputs possibly unseen label combinations, addressing the limitation of support inference. 

$\bullet$ show that the most elementary multi-label method  --- binary classifier for each label independently --- can be made almost as accurate as the most sophisticated methods, by employing the support inference and the GFM predictor during prediction time (post-training).

$\bullet$ reveal that the effectiveness of the pair-wise CRF model is likely not due to its ability to model pair-wise label dependencies, but mostly due to the support inference used in its implementation. 

$\bullet$ apply the proposed regularization techniques to several classifiers and achieve consistent improvements and state-of-the-art performance on many multi-label text datasets.

$\bullet$ code publicly available at \url{https://github.com/cheng-li/pyramid}.


\section{Multi-label Classifiers}
\label{sec:model}

 \begin{figure*}[t]
\centering
\includegraphics[width=1.8\columnwidth]{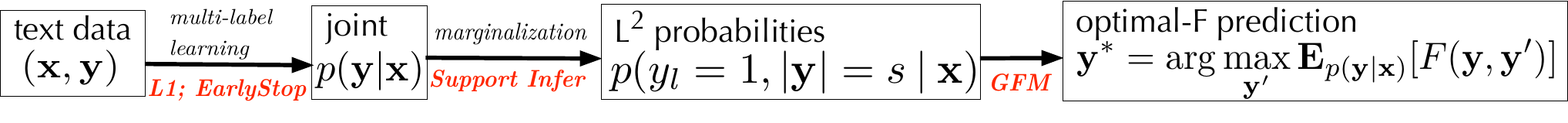}
\vspace{-1ex}
\caption{Our proposed pipeline with 4 regularization steps : L1 and EarlyStop (training); GFM and Support Inference (prediction). All learners also use L2 regularization.}
\label{fig:pipeline}
\end{figure*}

We briefly review several probabilistic multi-label classifiers that we shall use in the study.  For each method, we describe its probabilistic formulation $p(\mathbf{y}|\mathbf{x})$ and its MAP inference method, which gives $\arg\max_{\mathbf{y}} p(\mathbf{y}|\mathbf{x})$. We emphasize that MAP inference, although commonly used as the default choice, only provides the optimal prediction for the exact-set accuracy measure, but not for the F1 measure \cite{waegeman2014bayes, dembczynski2012label}.  In later sections, we will compare MAP prediction with the F1 optimal prediction. All these classifiers, except CRF, employ logistic regression as base learners, following the common practice in text classification.
 


\textbf { Binary Relevance  (BR)} \cite{tsoumakas2006multi} assumes that all labels are independent and thus the joint over all labels is a product of marginals
\[
p(\mathbf{y}|\mathbf{x})=\prod_{\ell=1}^{L}p(y_l|\mathbf{x})
\]

This independence assumption simplifies both training and prediction. During training, $L$ binary classifiers (logistic regressions) are trained, one for each label. The MAP inference simply predicts each label independently, based on the marginals.


\textbf { Probabilistic Classifier Chain  (PCC)} \cite{read2011classifier} decomposes the joint density estimator into a product of conditionals using the chain rule:
\[ p(\mathbf{y}|\mathbf{x})
=p(y_1|\mathbf{x})p(y_2|\mathbf{x},y_1)\cdots p(y_L|\mathbf{x},y_1,..,y_{L-1})\]
During training, one logistic regression is trained for each label based on both features and all previous labels; it models label dependency, but the pre-set order of labels in the decomposition is critical. The exact MAP inference is generally intractable. Beam search is often used as an approximate MAP inference procedure \cite{kumar2013beam}.

\textbf { Pair-wise Conditional Random Field  (CRF)}  \cite{ghamrawi2005collective} defines potential functions for each feature-label pair and each label-label pair and builds a log-linear model. 
\begin{align*}
p(\mathbf{y}|\mathbf{x})=&\frac{1}{Z(\mathbf{x})}\exp\{\sum_{l=1}^L\sum_{d=1}^D w_{ld}x_d\mathbbm{1}[y_l=1]\\
+&\sum_{l=1}^L\sum_{m=1}^L(w_{lm1}\mathbbm{1}[y_l=0,y_m=0]\\
+&w_{lm2}\mathbbm{1}[y_l=0,y_m=1]\\
+&w_{lm3}\mathbbm{1}[y_l=1,y_m=0]\\
+&w_{lm4}\mathbbm{1}[y_l=1,y_m=1])
\}
\label{eq:crf}
\end{align*} 
where $Z(\mathbf{x})$ is the normalization constant, and all $w$ are learned weights. All weights are trained jointly using maximum likelihood estimation. Both computing the partition function $Z(\mathbf{x})$  and the exact MAP inference are intractable, as an exponential number of label combinations are involved.  \cite{ghamrawi2005collective} suggests to use support inference to solve the intractability issues: the idea is to restrict $\mathbf{y}$ values only to label combinations observed in the training set when computing scores and probabilities. While the authors propose support inference only as an approximation, we think it is actually the main reason CRF works well: support inference adds strong regularization and dependency effects; the praised CRF design of pairwise dependencies has a smaller contribution on performance (can even be skipped) once the support inference is applied, at least on these datasets, as we shall show in our experiments.

\textbf {Conditional Bernoulli Mixtures  (CBM)} \cite{li2016conditional} represent the joint as a mixture of $K$ components, each with independent label classifiers
\[
p(\mathbf{y}|\mathbf{x}) =\sum_{k=1}^K \pi(z=k|\mathbf{x})  \prod_{{\ell}=1}^L b(y_{\ell}|\mathbf{x},z=k)
\]
The multi-class classifier (a multi-nomial logistic regression) $\pi$ decides the mixing coefficient for each mixture component. Inside each component, the joint is factorized into marginals, estimated by $L$ binary classifiers $b$ (binary logistic regressions). Both the multi-class classifier and the binary classifiers are trained jointly by EM algorithm.  The exact MAP can be computed efficiently using dynamic programming.




\section{Regularization for text data}
We apply two independent regularizations: during training we regularize the basic learner for feature selection and model size, using the  ``Elastic-net'' penalty and an ``Early Stopping'' criteria. 

During prediction, we apply the  ``GFM'' predictor to obtain an optimal instance-F prediction, and also the ``Support Inference'' which limits the prediction and some calculations to the label combinations observed in training set. 
These two work well together in that the ``Support Inference'' forces label dependencies in the model (even for independent classifiers), while the ``GFM'' beside optimizing F1, it allows for new label combinations to be predicted.

\subsection{Training Regularization: Model Complexity}
{\bf L1 Regularization.}
We take for granted that all base learners are L2 regularized (not debated in this paper). L2 regularization spreads weights across all correlated features; this makes the model more robust during testing. However, L2 does not perform any feature selection.
High capacity multi-label classifiers trained on high dimensional text features without careful feature selection can lead to severe over-fitting. One remedy for the high dimensionality is to apply the L1 regularization (LASSO) or the combinations of L1 and L2, which is also called Elastic-net \cite{friedman2010regularization}. L1 regularization shrinks some feature weights to zero  and thus can completely eliminate irrelevant features from the model, thus performing  powerful label-driven feature selection. However, L1 alone can pick one out of many highly correlated features, possibly hurting the generalization --- so L2 is necessary. Elastic-net regularization, with the form $\lambda\{\alpha ||w||_1+(1-\alpha)||w||_2^2\}$,  combines both L1 and L2 regularization and can get the best of both worlds, when properly tuned. Both the overall strength $\lambda$ and the relative ratio $\alpha$ can be tuned on a validation set.


For all multi-label methods except CRF, we implement elasic-net penalty to their linear base classifiers (i.e. binary and multi-nominal logistic regressions), using the coordinate descent training algorithm described in \cite{friedman2010regularization, yuan2012improved}. The elastic-net regularized multinomial Logistic Regression with $C$ classes (which includes binary logistic regression as a special case) is trained by minimizing the loss
\[
\frac{1}{N}\sum_{i=1}^{N}\log p_{g_i}(\mathbf{x}_i)-\lambda\sum_{c=1}^C [(1-\alpha)||w_c||_2^2+\alpha||w_c||_1],
\]
where
\[
p_c(x)=p(g=c| \mathbf{x})=\frac{e^{w_{0 c}+\mathbf{x}^T w_c}}{\sum_{c=1}^C e^{w_{0 c}+\mathbf{x}^T w_c}}.
\]
Regularizing CRF with elastic-net penalty is considerably more complicated \cite{lavergne2010practical}, and not considered in this paper (only L2 regularization is employed for CRF).



{\bf Early Stopping.}
It is often the case that even equipped with L1 and L2 regularization, the classifier can still over-fit the training data if the training algorithm is left to run until full convergence. Early stopping seeks to find the best termination point by monitoring the validation performance. Early stopping can be seen as a more general and  adaptive regularizer \cite{goodfellow2016deep}. So we also apply early stopping to all multi-label methods.

\subsection{Prediction Regularization: Label Structure}
\label{sec:pred}
The regularization techniques just described can potentially provide a good probabilistic joint model $p(\mathbf{y}|\mathbf{x})$. Then the next question is how to use this joint to make an accurate prediction for a given test instance $\mathbf{x}$. By ``accurate'', we mean a prediction that gives high F1 score when compared with the ground truth. Of course, the ground truth is unknown during prediction time, so we use Bayes optimal prediction to find the $\mathbf{y^*}$ that gives the highest expected F1 score:
\[
\mathbf{y^*}=\arg\max_\mathbf{y'} \mathbf{E}_{\mathbf{y}\sim p(\mathbf{y}|\mathbf{x})} [F(\mathbf{y},\mathbf{y'})]=\arg\max_\mathbf{y'} \sum_{\mathbf{y}} p(\mathbf{y}|\mathbf{x})\cdot F(\mathbf{y},\mathbf{y'})
\]

{\bf GFM: optimal prediction for F measure.}
The General F-Measure Maximizer (\textbf{GFM}) \cite{waegeman2014bayes} algorithm (see Algorithm 1) is an exact and efficient algorithm for computing $\mathbf{y^*}$.


However, the GFM algorithm does not work directly with a joint estimation  $p(\mathbf{y}|\mathbf{x})$, but rather, some marginal distributions of the form 
\[
p(y_l=1,|\mathbf{y}|=s\:|\:\mathbf{x}),\:\:\forall l, s\in\{1,...,L\},
\]
where $|\mathbf{y}|$ stands for the number of relevant labels in $\mathbf{y}$. This formula can be read as, for example, ``the probability of the given document having $s=5$ relevant labels and \texttt{election} is one of them''. There are (no more than) $L^2$ probabilities in this form, per instance.

\begin{algorithm}[h]
\caption{
General F-Measure Maximizer : given  $L^2$ probabilities of the form
 $r_{ls}=p(y_l=1,|\mathbf{y}|=s\:|\:\mathbf{x}); \forall l,s \in\{1,...,L\}$
  finds the $h(\mathbf{x})$ in $O(L^3$) time.
}
\begin{algorithmic}[1]
\STATE {\bfseries Input:} $p(\mathbf{y}=\mathbf{0}|\mathbf{x})$ and $L$ by $L$ Matrix $P$ with elements $p_{ls}=p(y_{\ell}=1,|\mathbf{y}|=s\:|\:\mathbf{x})$
\STATE Define  $L$ by $L$ Matrix $W$ with elements $w_{sk}=\frac{2}{s+k}$
\STATE Compute $L$ by $L$ Matrix $\Delta=PW$
\FOR {$s=1,2,...,L$}
\STATE The best prediction $\mathbf{y}^s$ with $s$ labels is given by including the $s$ labels $l$ with the highest $\Delta_{ls}$ 
\STATE Compute the expected F measure $\mathbf{E}[F(\mathbf{y},\mathbf{y}^s)]=\sum_{\ell=1}^L y^s_l \Delta_{\ell s}$
\ENDFOR
\STATE The expected F measure for the empty prediction $y^0=\mathbf{0}$ is $\mathbf{E}[F(\mathbf{y},\mathbf{y}^0)]=p(\mathbf{y}=\mathbf{0}|\mathbf{x})$.
\STATE {\bfseries Output:} optimal $\mathbf{y^*}=\argmax_{0\leq s\leq L}\mathbf{E}[F(\mathbf{y},\mathbf{y}^s)]$
\end{algorithmic}
\label{algorithm:prediction}
\end{algorithm}

\textbf{The LSF baseline.} It may appear that if the end goal is to produce these $L^2$ probabilities as input to the GFM algorithm, it should be more straight foreword to estimate these $L^2$ probabilities directly rather than going through a joint estimation $p(\mathbf{y}|\mathbf{x})$  first, which by itself is quite challenging (the joint evolves $2^L$ probabilities in general). In fact, it is conceptually not hard to derive an algorithm (we name it {\bf LSF}) to estimate the $L^2$ marginals directly, as suggested in 
\cite{waegeman2014bayes}. We can estimate  each $p(y_l=1,|\mathbf{y}|=s\:|\:\mathbf{x})$ using a binary logistic regression. Another option is to estimate $p(y_l=1|\mathbf{x})$ with a binary logistic regression, and then $p(|\mathbf{y}|=s\:|\:\mathbf{x}, y_l=1)$ with another multinomial logistic regression and then multiply their probabilities. However, in practice, we observe that this direct estimation approach does not perform well. Our speculation is that directly predicting the number of relevant labels $|\mathbf{y}|$ by a classifier is a very hard and unnatural task.  Each category here ("matching $s$ labels") does not have a fixed meaning, like \texttt{sport} or \texttt{economy} do in typical topical classification. As ``matching 2 labels'' can mean many different things, for example (\texttt{sport}, \texttt{basketball}) but also (\texttt{business}, \texttt{industry}), it is hard to establish a relation between "matching 2 labels" and feature representation. This is especially so for linear models like logistic regression, where probabilities are monotonic functions of scores, which in turn are monotonic functions of features. 



\textbf{Support Inference. }
\label{sec:support}
Estimating the joint $p(\mathbf{y}|\mathbf{x})$ is a more natural formulation and is what most of the methods do. Now we discuss how to make GFM work with the existing joint estimator. First, it is generally impossible to compute the required $L^2$ marginal probabilities analytically from the joint estimation.
\cite{waegeman2014bayes} proposed to sample from the joint and then compute the required probabilities based on the samples. Sampling works best when it takes advantage fo the classifier structuure: For BR each label can be sampled independently; in PCC one can sample labels one by one, using  ancestral sampling; in CBM one can first sample a component, and from the component sample each label independently. However sampling can be very inefficient for large $L$ and in particular low-confidence ("flat") joint that allows probability mass to spread over many label combinations.

\begin{table*}[!t]
\centering
\begin{tabular}{|c|c|c|c|c|c|c|c|c|c|}
\hline
\textbf{}                & {MEDICAL} 	& {BIBTEX} 		&{IMDB} 	& {OHSUMED} 		& {ENRON} 		& {RCV1} 	&{TMC} 		&{WISE} 		&{WIPO} \\ \hline
\textbf{domain}          & medical          & bkmark        & genre      & medical        	& email         	& news      & reports 		& articles 		& patent         \\ \hline
\textbf{labels}          & 45               & 159           & 27         & 23             	& 53          	& 101       & 22      		& 203 		& 188          \\ \hline
\textbf{label sets}   	 & 90               & 2,058          & 2,122     & 1,042       	 	& 653    		& 494       & 1,172   		& 3,536 		& 155           \\ \hline
\textbf{features}        & 679             & 1,836          & 27,228     & 16,344     		& 21,190   		& 47,236    & 49,060   		& 301,561 		& 74,435           \\ \hline
\textbf{instances}       & 978              & 7,395          & 34,157     & 13,929      	 	& 1,702           & 6,000     & 28,596  		& 64,857 		& 1,710           \\ \hline
\textbf{cardinality}     & 1.25             & 2.40           & 2.52       & 1.66          	& 3.38             & 3.23     & 2.16    		& 1.45 		& 4.00         \\ \hline
\textbf{inst/label} 	 & 27               & 112            & 2537        & 1007             	& 108              & 188      & 2,805     	& 463 		& 36         \\ \hline
\end{tabular}
\label{table:data}
\caption{Datasets characteristics; cardinality is the average number of labels of the instances and instances/label denotes the average number of training instances of the labels. The IMDB dataset is put together by us from different sources, the other ones are publicly available.}
\end{table*}


Support Inference not only allows us to infer the required marginals from the joint, but also provides some additional regularization effect on the label structures. We say a combination $\mathbf{y}$  is a support combination if it appears at least once in the training set. For each support combination $\mathbf{y}$, we compute $p(\mathbf{y}|\mathbf{x})$ using the trained classifier. We then re-normalize all support combination probabilities to make them sum up to 1 (this is only for conceptual demonstration purpose; in practice, this step can be skipped as the constant factor does not affect the final GFM prediction). By doing so, we have put all the probability mass on only support combinations. Using these limited number (usually no more than a few thousands, see Table~1) of support combinations and their associated probabilities, we can then easily compute the required marginals.


To illustrate how support inference could possibly help regularizing label structures and modeling label dependencies, let us consider a toy example. Suppose we have a multi-class problem with 3 labels (which can be seen as a special multi-label task). Then the only valid combinations are the singleton sets \{1\}, \{2\} and \{3\}. Binary relevance  may assign substantial probabilities to  the invalid set \{1, 2\} or the empty set $\emptyset$. Pairwise CRF can learn that no two labels should co-occur and thus can largely avoid \{1, 2\}, but it cannot learn that exactly one out of three labels has to be picked, since this rule involves higher order dependencies; as a result, pairwise CRF still considers the empty label set $\mathbf{y}=\emptyset$ as a valid candidate. But when support inference is employed, both binary relevance and CRF will assign zero posterior probabilities to these invalid empty sets. Thus support inference can be seen as an infinite strong prior which regularizes the posterior to only consider observed label combinations. In practice, this prediction-time regularization often allows a classifier to capture label dependencies that the classifier itself did not capture during training.

Admittedly, support inference has the limitation in that no new label combination will be considered \emph{during marginalization}. The datasets used in this study have only few new label combinations in the test set; but even if the data would have such new label combinations, this support-inference limitation is largely mitigated by GFM: GFM can output a $\mathbf{y^*}$ that maximizes the expected F1 even when $\mathbf{y^*}$ is not in the support and thus $p(\mathbf{y^*}|\mathbf{x})=0$. Using the example above and assuming support inference gives $p(\{1\}|\mathbf{x})=0.5$, $p(\{2\}|\mathbf{x})=0.4$ and $p(\{3\}|\mathbf{x})=0.1$ and $p(\mathbf{y}|\mathbf{x})=0$ for all other $\mathbf{y}$. In this case, the F1 optimal prediction is $\mathbf{y^*}=\{1, 2\}$, with expected F1  $0.5\times\frac{2}{3}+0.4\times\frac{2}{3}+0.1\times 0=0.6$. As a comparison, the joint mode $\mathbf{y}=\{1\}$ (which maximizes the set accuracy) has an expected F1 of only $0.5\times 1 +0.4\times 0+0.1\times 0=0.5$. Thus support inference provides a regularized probability estimation by only assigning probability mass to observed combinations, and GFM takes this regularized probability estimation as the input and outputs F optimal prediction, which could potentially be an unobserved label combination. We observe this strategy to work remarkably well for many classifiers and datasets.




%





\section {Experimental Results \& Analysis}

In this section, we empirically verify the effectiveness of the proposed regularization techniques L1+L2, early stopping, support inference and GFM for 5 multi-label methods on 9 multi-label text datasets.

\subsection{Datasets and Experiment Setup}
\label{sec:dataset}
The 9 multi-label text datasets used are shown in Table 1. We adopt the given train/test split whenever it is provided; otherwise we use a random 20\% of the data as the test set. We further split 20\% data from the training data as the validation set. Hyper parameter tuning for all algorithms is done on the validation set and F1 metric on the test set is reported.
For methods involving random initializations or sampling, reported results are averaged over 3 runs. 
When we apply L1 regularization, we use L1 penalty together with the basic L2 penalty in the elastic-net form $\lambda\{\alpha ||w||_1+(1-\alpha)||w||_2^2\}$, and we tune the overall strength $\lambda$ and the L1 ratio $\alpha$.  When L1 penalty is not included, we only keep L2 penalty by setting $\alpha=0$ and we only tune $\lambda$. For early stopping, we tune the optimal number of training iterations on the validation set. When early stopping is not applied, the training is left running until full convergence. Support inference and GFM do not contain any tunable hyper parameters. When GFM is not used, we perform MAP inference instead. The number of mixture components in CBM is fixed as 20.




\subsection{L1 Regularization and Early Stopping}

\begin{table} \centering 

\begin{tabular}{cc|ccccc} 
\\[-1.8ex]\hline 
\hline \\[-1.8ex] 
\multicolumn{1}{c}{Data} & \multicolumn{1}{c}{Model} & \multicolumn{1}{c}{No REG} & \multicolumn{1}{c}{S+G} & \multicolumn{1}{c}{L+S+G} & \multicolumn{1}{c}{E+S+G} & \multicolumn{1}{c}{All4 REG} \\ 
\hline \\[-1.8ex] 
\multirow{4}{*}{\rotatebox{90}{MEDICAL}}
		& BR  &  68.8 & 80.5 &  81.1 &  80.3 & \bf{81.1} \\
 		& LSF &	 - & 81.3 &  81.6 &  81.4 & \bf{82.1}  \\
		& PCC & 73.6 &  78.3 & 80.6 & 78.3 & \bf{80.8} \\
		& CBM &  79.2 & 80.1 & \bf{82.6}* & 81.3 & \bf{82.6}* \\

\hline \\[-1.8ex] 
\multirow{4}{*}{\rotatebox{90}{BIBTEX}}
		& BR & 37.8 & 44.5 & \bf{45.5} & 45.4 & 45.4 \\
		& LSF &	 - & 42.5 &  43.3 &  43.2 & \bf{43.9}  \\
		& PCC & 37.6 & 45.3 & 45.8 & 45.3 & \bf{47.3} \\
		& CBM & 44.0 & 45.9 & 47.3 & 46.5 & \bf{49.5}* \\

\hline \\[-1.8ex]
\multirow{4}{*}{\rotatebox{90}{IMDB}} 
	& BR &  59.4 & 61.8 & \bf{61.4} & 63.1 & \bf{61.4} \\
	& LSF &	 - & 59.9 &  \bf{60.0} &  59.4 & 59.8  \\
	& PCC & 59.6 & \bf{63.9} & 62.8 &  \bf{63.9} & 62.8 \\
	& CBM & 61.6 & 65.1 & 65.0 & \bf{65.4}* & 65.2 \\

\hline \\[-1.8ex]
\multirow{4}{*}{\rotatebox{90}{OHSUMED}} 
		& BR &  60.2 &  67.9 & 68.8 & 68.3 & \bf{69.1} \\
		& LSF &	 - & 64.2 &  64.5  &  64.2 & \bf{65.0} \\
		& PCC & 62.5 & 70.1 & 69.2 & 70.1 &  \bf{70.4}\\
		& CBM & 68.7 & 70.3 & 71.1 & 70.1 &  \bf{71.7}* \\

\hline \\[-1.8ex]
\multirow{4}{*}{\rotatebox{90}{ENRON}} 
		 & BR & 57.9 & \bf{61.1} & \bf{61.1} & 60.4 & \bf{61.1} \\
		& LSF &	 - & \bf{58.0} &  55.2 &  57.8 & 57.6  \\
		& PCC & 60.1 & 61.0 & \bf{61.5} & 61.0 & \bf{61.5}\\
		& CBM & 58.2 &  59.8 & 60.9 & 61.1 & \bf{62.5}* \\

\hline \\[-1.8ex]
\multirow{4}{*}{\rotatebox{90}{RCV1}} 
		 & BR & 72.1 &  73.7 & \bf{75.1} & 73.8 &  \bf{75.1} \\
		& LSF &	 - & 73.4 &  73.5 &  73.4 & \bf{73.6}  \\
		& PCC & 71.0 &  73.6 & \bf{74.1} & 73.6 & \bf{74.1}\\
		& CBM & 76.6 &  77.3 & 78.9  & 77.7 & \bf{79.2}*\\

\hline \\[-1.8ex]
\multirow{4}{*}{\rotatebox{90}{TMC}} 
 		& BR & 61.6 &  \bf{63.2} & 63.1 & 62.4 & 63.1 \\
		& LSF &	 - & 58.9 &  55.9 &  \bf{60.0} & 57.6  \\
		& PCC & 61.9 & \bf{63.4} & 63.2 & \bf{63.4} & 63.2\\
		& CBM & 59.8 &  61.7 & 62.0 & \bf{63.8}* & \bf{63.8}*\\

\hline \\[-1.8ex]
\multirow{4}{*}{\rotatebox{90}{WISE}} 
 		& BR & 68.0 & 77.3 &  \bf{79.6} & 78.6 &  79.3 \\
		& LSF &	 - & 75.9 &  76.5 &  76.3 & \bf{76.7}  \\
		& PCC & 70.7 & 76.0 & 77.2 & 76.1 & \bf{78.0}\\
		& CBM & 77.9 & 78.6 & \bf{80.4}* & 78.7 & 80.3 \\

\hline \\[-1.8ex]
\multirow{4}{*}{\rotatebox{90}{WIPO}} 
		& BR & 63.4 & 71.2 & 73.1 & 71.9 &  \bf{74.0} \\
		& LSF &	 - & 65.0 &  71.0 & 65.2  & \bf{71.1}  \\
		& PCC & 68.8 &  71.5 & 71.2 & 70.7 & \bf{72.3}\\
		& CBM & 63.0 &   70.8  &  73.9 &  71.8 & \bf{74.3}* \\

\hline \\[-1.8ex] 
\end{tabular} 
  \caption{performance (F1 on test) w/ and w/out Training Regularization L1, Early Stop. L=L1; E=Early Stopping; S=Support Inference; G=GFM prediction. CRF excluded as not L1 regularized during training. bold=best in row, *=best in dataset} 
    \label{trainTable} 
  \vspace{-6ex}
\end{table}

First we analyze the regularization effects of L1 penalty and early stopping during training. The experiment results are summarized in in Table~\ref{trainTable}. The ``No REG'' column does not use any of the proposed training or prediction regularization techniques (Elastic-net, early stopping, support inference, GFM). ``No REG'' follows the convention that uses only L2 penalty to regularize logistic regression learners, trains each model until full convergence, and perform MAP inference during prediction. This column serves as a baseline. All other columns have the prediction strategy fixed as support inference + GFM, and only vary the training regularizations. The letters ``L, E, S, G'' in the table stand for L1, early stopping, support inference and GFM, respectively, and ``All4'' stands for ``L+E+S+G'', i.e., using all four techniques. We did not implement L1 regularization for CRF, and thus do not include CRF in this table. Comparing the column ``L+S+G''  with the column ``S+G'', we can see the difference due to L1. The results show that how L1 influences the results depends more on the  datasets, and less on the classifiers. The 
performance of all classifiers generally improves on MEDICAL, BIBTEX, OHSUMED, RCV1, WISE, and WIPO, but not on IMDB, ENRON and TMC. 

\begin{figure}[h]
\centering
\includegraphics[width=.49\columnwidth]{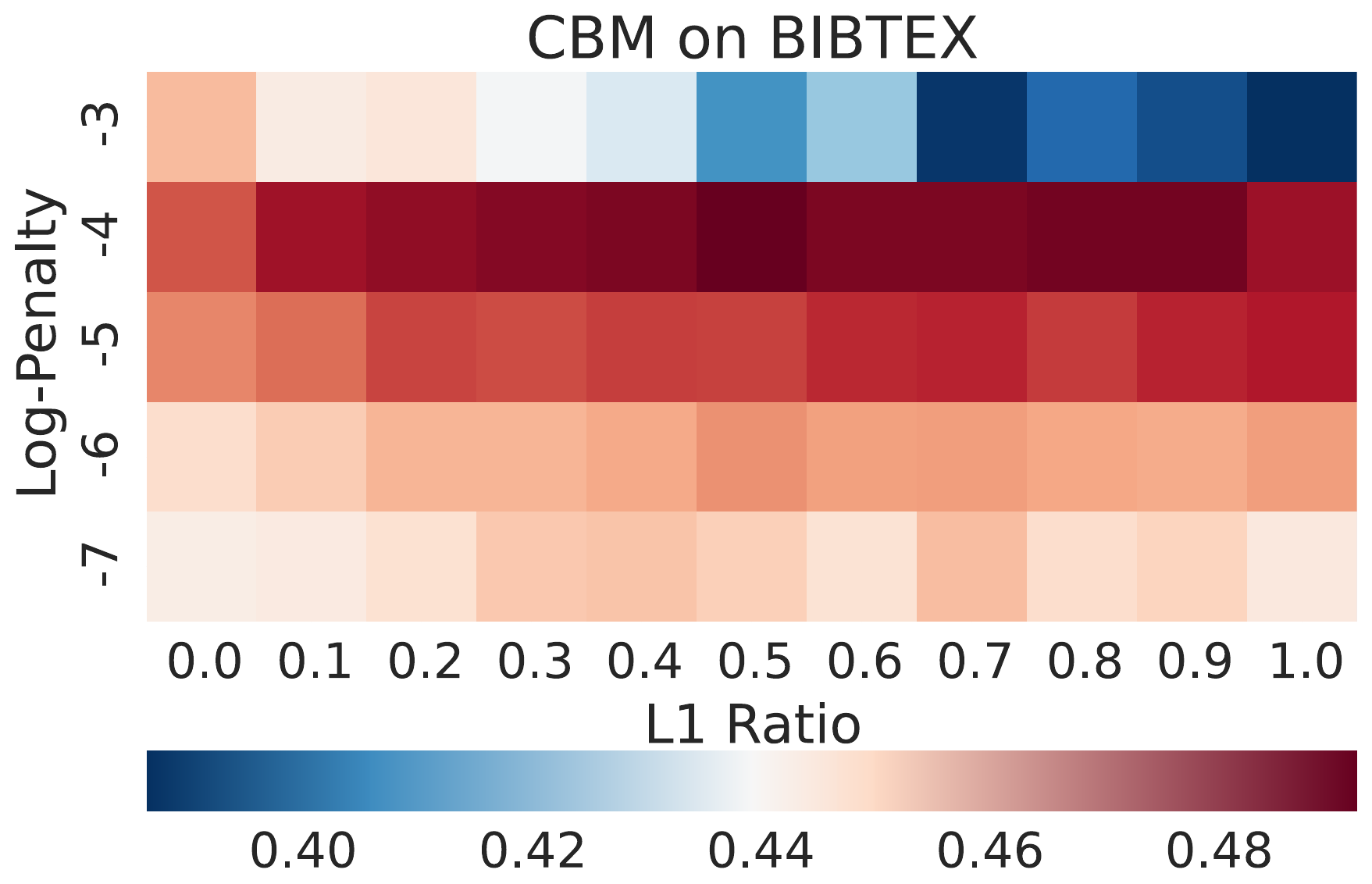}
\includegraphics[width=.49\columnwidth]{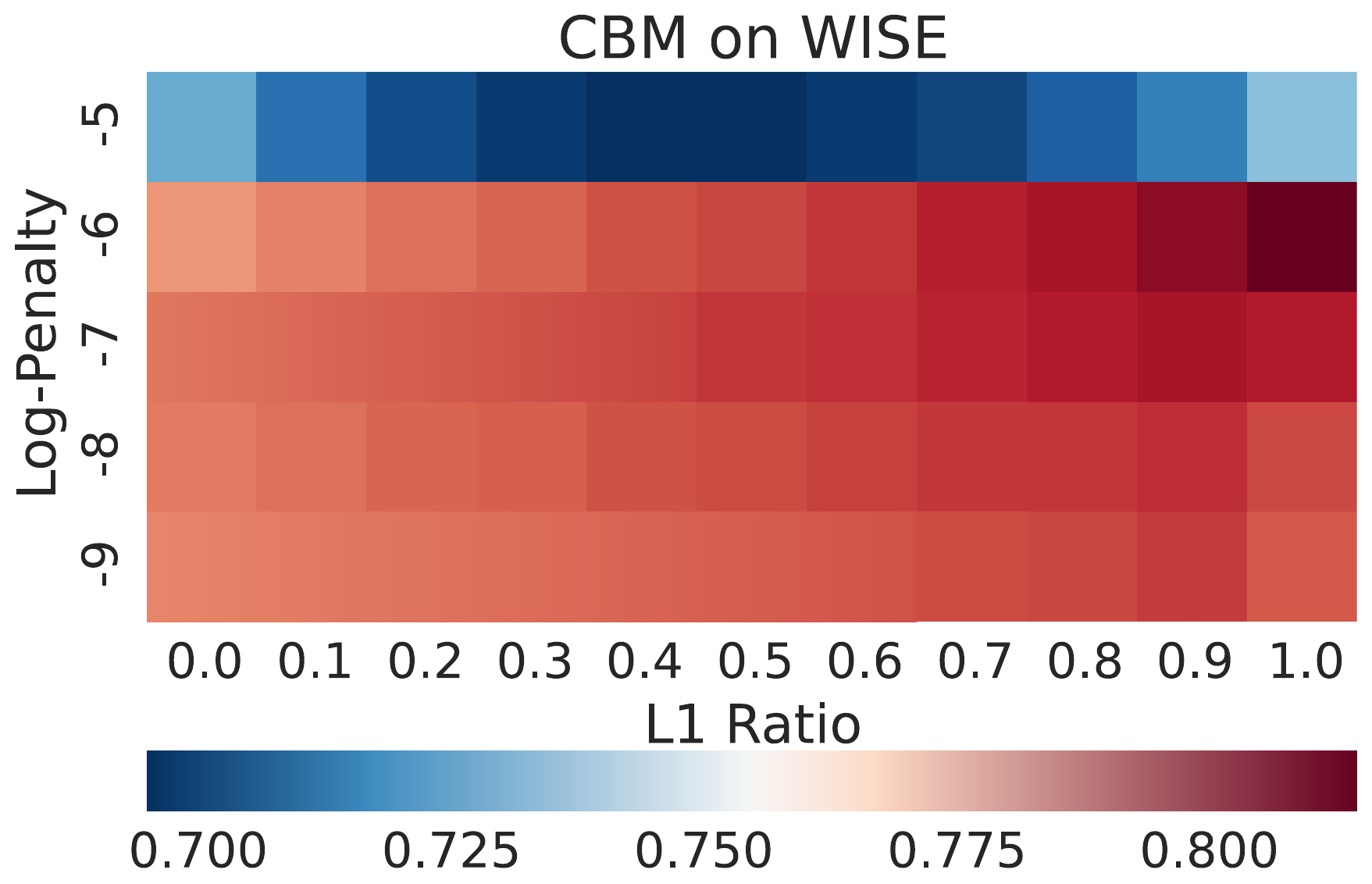}\\
\vspace{3ex}
\includegraphics[width=.49\columnwidth]{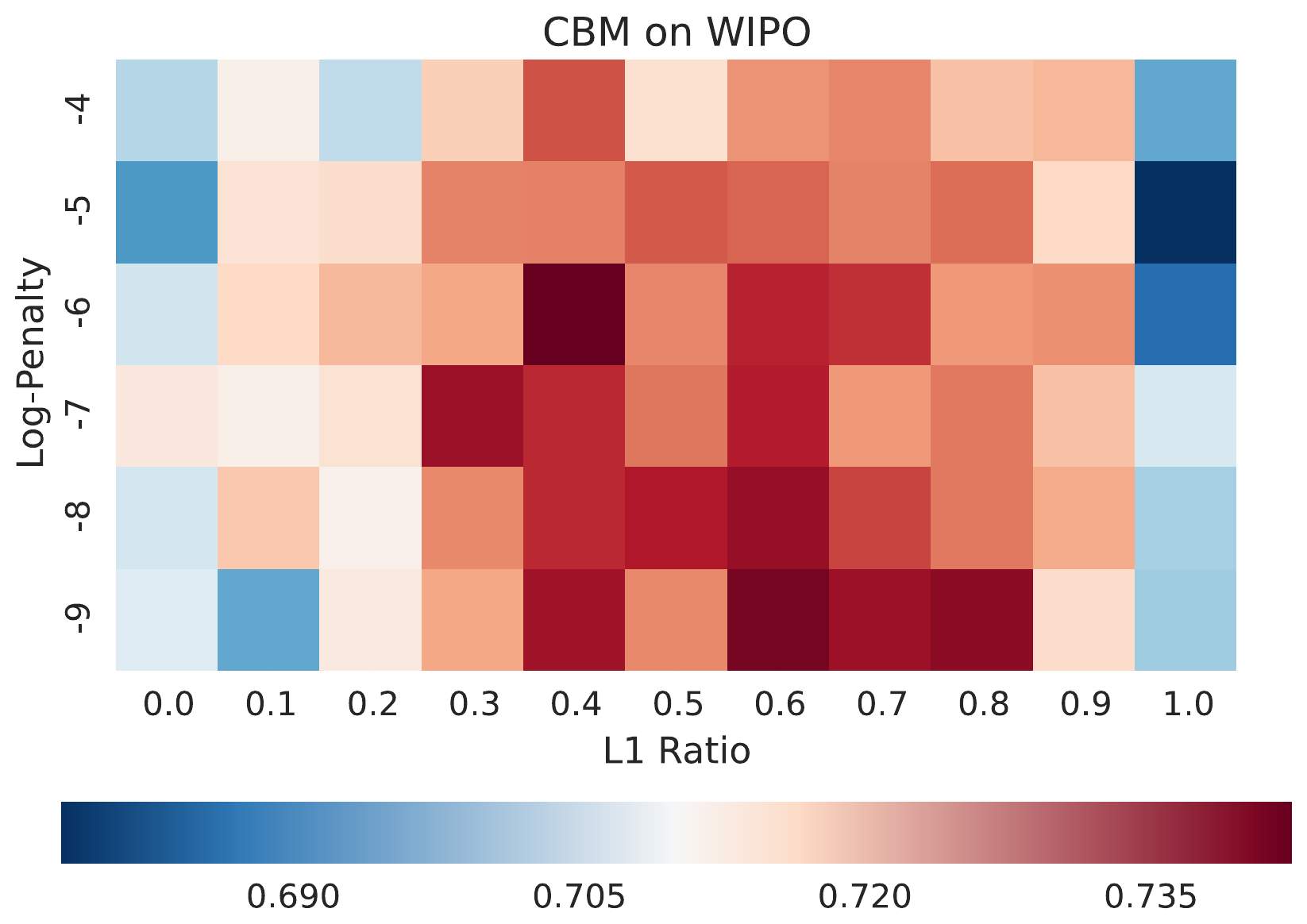}
\includegraphics[width=.49\columnwidth]{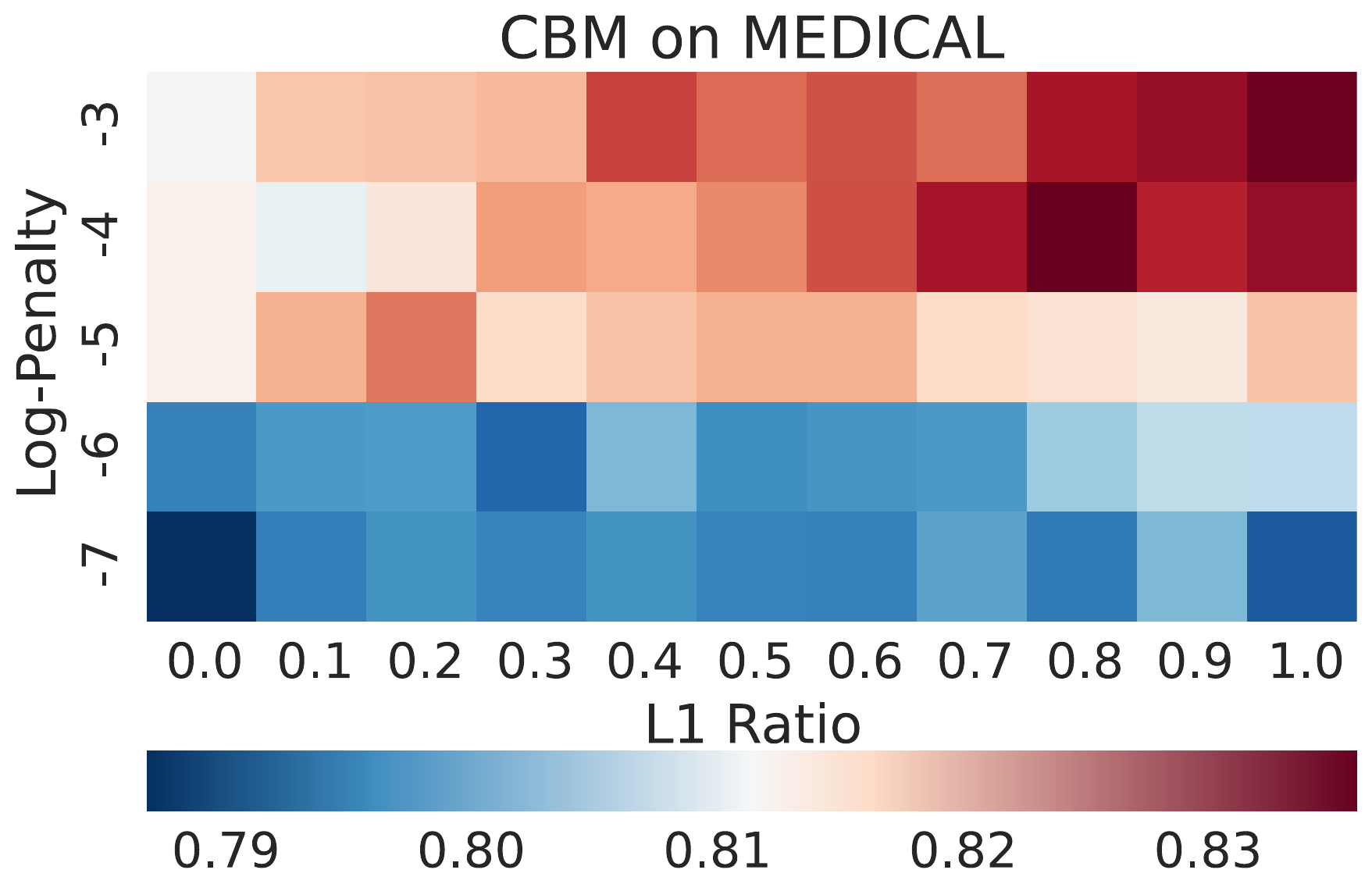}\\
\vspace{3ex}
\includegraphics[width=.49\columnwidth]{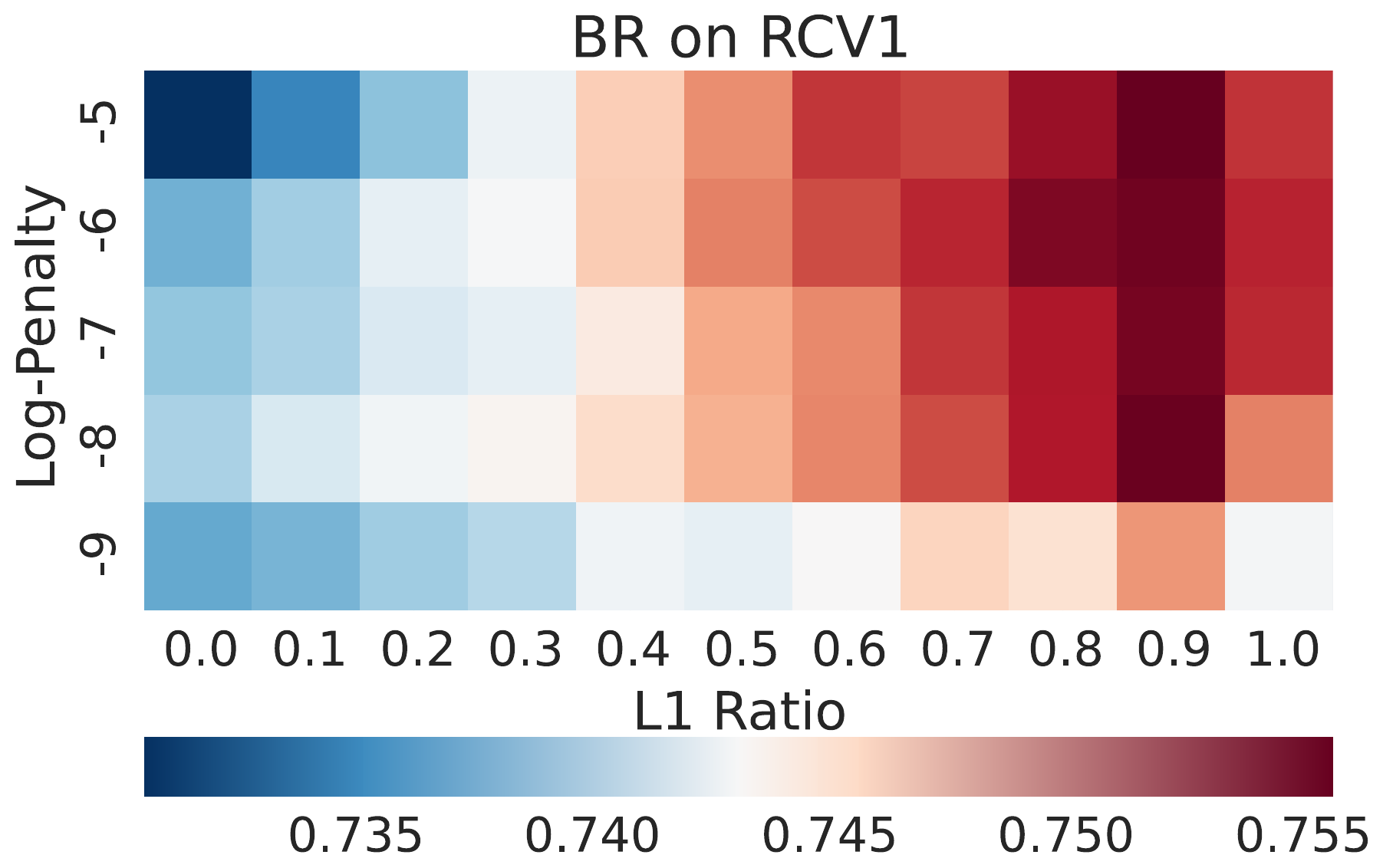}
\includegraphics[width=.49\columnwidth]{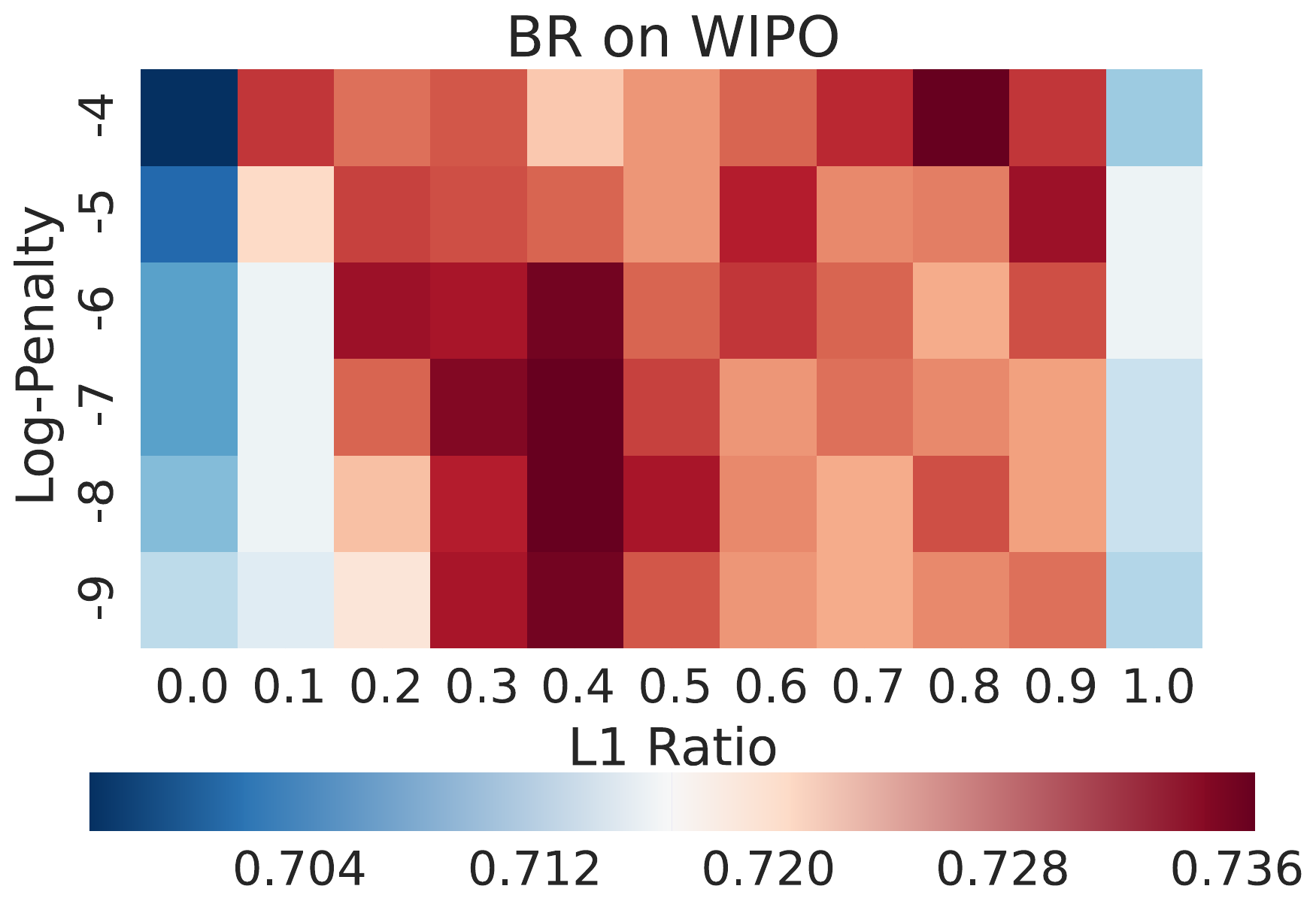}
\caption{Test F1 versus the penalty strength and the L1 ratio. The penalty strength is on the log scale.}
\label{fig:heatmapCons}
\end{figure}

One can imagine that each dataset has some intrinsic properties such as the number of relevant documents per label, the diversity of the topics, the total number of documents and the total number of features, that dictate how many features have to be used in order to explain the given labels/topics well  and how many model parameters can be reliably estimated based on the given dataset size, and these factors in turn influence how much performance improvement L1 feature selection can bring in. The heap maps in Figure~\ref{fig:heatmapCons} shows how varying the overall regularization strength and the L1 ratio affects the test performance. One can see that the optimal amount of L1 penalty varies on different datasets. Overall introducing some L1 penalty leads to the performance improvement on 6 out of 9 datasets.

Apart from improving test performance, L1 also shrinks the model sizes massively. The model size is measured by the disk space the model occupies. Since the model weights are sparse when L1 is added, only non-zero weights are saved. Table~\ref{modelComplexity}  compares the sizes of models trained with only L2 penalty versus models trained with both L1 and L2 penalty. Generally, adding L1 shrinks models  to no more than 10\% of its original sizes. On some datasets, such as ENRON, RCV1, WISE and WIPO, the shrunk CBM models are only about 1\% of its original sizes. Interesting, if we look at the total number of features selected by the classifiers (union of features used in all base learners), the reduction in feature size is not as dramatic as the reduction in model size. This is in direct contrast with binary classification, where these two reductions mostly agree. By looking into the trained multi-label classifiers, we notice that although many features are relevant for some labels and are thus included in the classifiers, for each individual label, only a few features actually have non-zero weights. Thus although the union of all relevant features for all labels can be large, each label predictor can be a small model that includes a few features, and the entire multi-label classifier is therefore quite compact.







\begin{table} \centering 
 \resizebox{1.01\columnwidth}{!}{
\begin{tabular}{@{}c|cc|cc|cc}
\\[-1.8ex]\hline 
\hline \\[-1.8ex] 
\multirow{2}{*}{model} & \multicolumn{2}{c}{BR} & \multicolumn{2}{c}{LSF} & \multicolumn{2}{c}{CBM} \\

& {Features} & {Model} & {Features} & {Model} & {Features} & {Model} \\
& {selected} & {size} & {selected} & {size} & {selected} & {size} \\
\hline \\[-1.8ex]

MEDICAL & 60.08 & 14.89 &  75.40 & 12.62 & 81.15 &  6.67 \\
\hline \\[-1.8ex]
BIBTEX & 100.00 &  26.08 &   99.89 & 11.25 & 100.00 &   4.44  \\
\hline \\[-1.8ex]
IMDB & 66.41 &  20.5 &   100.00 & 7.06 &  99.54 &   9.57  \\
\hline \\[-1.8ex]
OHSUMED & 52.62 &  33.67 &   61.81 & 5.09 &  68.42  &   5.54  \\
\hline \\[-1.8ex]
ENRON & 31.81 &  7.5 &   26.24 &2.41  &  50.24 &   1.54  \\
\hline \\[-1.8ex]
RCV1 & 69.99 &  12.29  &   79.41  &7.98  &  77.29 &   1.52  \\
\hline \\[-1.8ex]
TMC & 37.87 &  13.84 &   40.54 & 2.82 &  78.22  &    2.84  \\
\hline \\[-1.8ex]
WISE & 14.15 &  1.12 &   20.18 &0.48 &  24.41  &    0.33  \\
\hline \\[-1.8ex]
WIPO & 41.64 & 2.07 &   38.75 &2.02 &  76.57   &    1.54  \\
\hline \\[-1.8ex] 
\end{tabular} 
}
  \caption{Model Size and Feature Used after adding L1 in CBM, LSF, BR (as percentages of the original sizes)}
  \label{modelComplexity}  
\end{table}
To understand the impact of early stopping, we compare the ``S+G'' column with the ``E+S+G'' column. We observe that unlike L1, which is more dataset dependent, early stopping is more algorithm dependent. Early stopping helps most for CBM, and less for other methods. CBM has higher capacities and is more likely to over-fit compared with other methods, and thus benefits more from early stopping. Comparing the last 4 columns of the table, it is clear that typically the best performance is achieved when both L1 and early stopping are employed.

\subsection{GFM prediction and Support Inference}
\label{sec:impactofGFM}

\begin{table} \centering 

\begin{tabular}{cc|ccccc} 
\\[-1.8ex]\hline 
\hline \\[-1.8ex] 
\multicolumn{1}{c}{Data} & \multicolumn{1}{c}{Model} & \multicolumn{1}{c}{No REG} & \multicolumn{1}{c}{L+E} & \multicolumn{1}{c}{L+E+S} & \multicolumn{1}{c}{L+E+G} & \multicolumn{1}{c}{All4 REG} \\ 
\hline \\[-1.8ex] 
\multirow{4}{*}{\rotatebox{90}{MEDICAL}}
		& BR  &  68.8 & 75.4 &  \bf{81.5} &  76.3 & 81.1 \\
 		& CRF &	 - & 77.8  &  77.8 &  - & \bf{79.4}  \\
		& PCC & 73.6 &  79.0 & 80.1 & 80.6 & \bf{80.8} \\
		& CBM &  79.2 & 82.2 & 82.3 & 79.6 & \bf{82.6}* \\

\hline \\[-1.8ex] 
\multirow{4}{*}{\rotatebox{90}{BIBTEX}}
		& BR & 37.8 & 39.8 & 44.4 & 40.2 & \bf{45.4} \\
		& CRF &	 - & 46.5 &  46.5 &  - & \bf{49.4}  \\
		& PCC & 37.4 & 39.5 & 45.0 & 40.1 & \bf{47.3} \\
		& CBM & 44.0 & 45.3 & 46.9 & 40.4 & \bf{49.5}* \\

\hline \\[-1.8ex]
\multirow{4}{*}{\rotatebox{90}{IMDB}} 
	& BR &  59.4 & 59.6 & 59.7 & 61.0 & \bf{61.4} \\
	& CRF &	 - & 63.0 &  63.0 &  - & \bf{66.6}*  \\
	& PCC & 59.6 & 60.1 & 60.2 &  61.5 & \bf{62.8} \\
	& CBM & 61.6 & 62.2 & 62.2 & 64.8 & \bf{65.2} \\

\hline \\[-1.8ex]
\multirow{4}{*}{\rotatebox{90}{OHSUMED}} 
		& BR &  60.2 &  63.6 & 68.0 & 64.3 & \bf{69.1} \\
		& CRF &	 - &  66.4 &  66.4  &  - & \bf{69.6}  \\
		& PCC & 62.5 &  64.7 & 68.4 & 65.8 &  \bf{70.4}\\
		& CBM & 68.7 & 69.5 & 70.3 & 65.4 &  \bf{71.7}* \\

\hline \\[-1.8ex]
\multirow{4}{*}{\rotatebox{90}{ENRON}} 
		 & BR & 57.9 & 59.8 & 59.3 & 60.4 & \bf{61.1} \\
		& CRF &	 - & 60.2 &  60.2 &  - & \bf{64.1}*  \\
		& PCC & 60.1 & 60.5 & 60.3 & 61.2 & \bf{61.5}\\
		& CBM & 58.2 & 58.5 & 60.9 & 59.2 & \bf{62.5} \\

\hline \\[-1.8ex]
\multirow{4}{*}{\rotatebox{90}{RCV1}} 
		 & BR & 72.1 &  73.8 & 74.6 & 74.9 &  \bf{75.1} \\
		& CRF &	 - & 74.4 &  74.4 &  - & \bf{75.8}  \\
		& PCC & 71.0 &  72.7 & 72.8 & \bf{74.3} & 74.1\\
		& CBM & 76.6 &  77.3 & 78.5  & 77.9 & \bf{79.2}*\\

\hline \\[-1.8ex]
\multirow{4}{*}{\rotatebox{90}{TMC}} 
 		& BR & 61.6 &  61.6 & 62.6 & 62.2 & \bf{63.1} \\
		& CRF &	 - & 62.0 &  62.0 &  - & \bf{64.7}*  \\
		& PCC & 61.9 & 62.0 & 62.4 & 62.7 & \bf{63.2}\\
		& CBM & 59.8 & 60.4 & 62.7 & 60.7 & \bf{63.8}\\

\hline \\[-1.8ex]
\multirow{4}{*}{\rotatebox{90}{WISE}} 
 		& BR & 68.0 & 72.8 &  79.0 & 73.0 &  \bf{79.3} \\
		& CRF &	 - & 77.7 &  77.7 &  - & \bf{79.0}  \\
		& PCC & 70.7 & 74.6 & 76.7 & 77.1 & \bf{78.0}\\
		& CBM & 77.9 & 79.8 & 79.8 & 73.6 & \bf{80.3}* \\

\hline \\[-1.8ex]
\multirow{4}{*}{\rotatebox{90}{WIPO}} 
		& BR & 63.4 & 69.5 & 73.2 & 70.0 &  \bf{74.0} \\
		& CRF &	 - & 70.3 &  70.3 &  - & \bf{72.2}  \\
		& PCC & 68.8 &  70.2 & 70.4 & 70.6 & \bf{72.3}\\
		& CBM & 63.0 &  69.6  & 72.5 & 70.3 & \bf{74.3}* \\

\hline \\[-1.8ex] 
\end{tabular} 

\raggedright\scriptsize\emph{Note: } {'-': Not available; ElasticNet for CRF is L2 regularization only.}
    \caption{performance (instance-F1 values on test) w/ and w/out Prediction Regularization:  Support Inference and GFM. L=L1; E=Early Stopping; S=Support Inference; G=GFM prediction. LSF excluded since it must use GFM and cannot use Support Inference. bold=best in row, *=best in dataset} 
  \label{predTable} 
  \vspace {-4ex}
\end{table}
We now compare the proposed support inference + GFM prediction strategy with other prediction strategies. Experiment results are summarized in Table~\ref{predTable}. LSF is designed to always work with GFM and the concept of support inference do not apply to LSF, so LSF is  not included in this table. As before, the ``No REG'' column does not use any of the proposed training time or prediction time regularizations and serves as a baseline. All other columns  have L1 and early stopping regularizations applied during model training and  only vary the prediction methods.  The ``L+E'' column uses MAP inference as described in Section~\ref{sec:model}. The ``All4'' column uses the proposed support inference + GFM prediction strategy. Comparing the ``L+E'' column with the ``All4'' column it is clear that support inference + GFM prediction performs better than MAP prediction for all methods on all datasets.

\begin{table*}[b] \centering 

\begin{tabular}{@{}cc|cccccccccc} 
\\[-1.8ex]\hline 
\hline \\[-1.8ex] 
\multicolumn{2}{c}{} & \multicolumn{1}{c}{MEDICAL} & \multicolumn{1}{c}{BIBTEX} & \multicolumn{1}{c}{IMDB} & \multicolumn{1}{c}{OHSUMED} & \multicolumn{1}{c}{ENRON} & \multicolumn{1}{c}{RCV1} & \multicolumn{1}{c}{TMC}& \multicolumn{1}{c}{WISE} & \multicolumn{1}{c}{WIPO}\\ \hline \\[-1.8ex] 
\multirow{2}{*}{CRF w/o pair label depend} & w/o GFM&  79.2 &  46.9 & 61.3 & 66.1 & 59.8 & 73.8 & 62.6 & 78.2 & 70.7\\
							 & w/ GFM  &  80.5 &  49.4 & 66.1 & 69.8 & 64.4 & 75.8 & 64.5 & 79.4 & 71.8 \\

\hline \\[-1.8ex] 
\multirow{2}{*}{CRF w/ pair label depend} & w/o GFM &  77.8 & 46.5  & 63.0 & 66.4 & 60.2 & 74.4 & 62.0 & 77.7 & 70.3 \\
 							& w/ GFM &  79.4 & 49.4 & 66.6 & 69.6 & 64.1 & 75.8 & 64.7 & 79.0 & 72.2 \\

\hline \\[-1.8ex] 
\end{tabular}
  \caption{Instance F1 compared bewteen CRF w/ and  w/o pairwise dependencies} 
  \label{crfCompare}   
\end{table*}
Next we break down the support inference + GFM prediction combination and test how each of them works separately. The ``L+E+S'' column only uses support inference but not GFM prediction. We use support inference to restrict the label combinations to those observed in the training set, and among them, we  pick the one with the highest probability. So this is basically MAP inference restricted on support combinations. Comparing the ``L+E'' column with the ``L+E+S'' column, we see that adding the support inference alone consistently improves the test performance (except for CRF, for which ``L+E'' is the same as ``L+E+S'', as described in Section~\ref{sec:model}). The improvement is most substantial on BR, which did not estimate any label dependencies during training, and is relatively small on CRF, PCC and CBM, which already estimated some label dependencies during training. This observation matches our analysis in Section~\ref{sec:pred} that support inference acts as a regularizer on the label structures -- it allows the classifier to take into account label dependencies during prediction even if the classifiers had not modeled any label dependencies during training.

It is known that if CRF as described in Section~\ref{sec:model} only contains label-feature interaction but not label-pair interaction, and if the partition function is computed exactly by summing overall all label combinations (conceptually), then the resulting model is mathematically equivalent to BR. In this case, theoretically there is no need to compute the partition function approximately using support combinations. So the authors in \cite{ghamrawi2005collective} only use support inference when CRF contains label pair interactions and thus the model does not factorizes and one has to somehow compute the partition function. Support inference there was deemed purely as an approximate inference procedure. Here our results show, perhaps surprisingly,  that support inference in fact helps on BR -- in other words, even if we could compute the CRF partition function exactly, it is still beneficial to compute it approximately, using support inference, which helps to capture label dependencies. Then it is  natural to raise a question regarding pair-wise CRF, which is one of the state-of-the-art multi-label classifiers: does its good performance come from its pair-wise label interaction terms, or from this support inference?  To test this idea, we removed the label pair terms in CRF. Table~\ref{crfCompare} shows that CRF without label pair terms do almost equally well as the one with label pair terms. This analysis is not to criticize the design of pair-wise CRF (in fact, our study has drawn lot of inspiration from the paper \cite{ghamrawi2005collective}), but rather, to shed lights on the effectiveness of CRF, and to bring to  researchers' attention of support inference as a simple yet powerful 	regularizer, besides its original role as an approximation.

The ``L+E+G'' column only uses GFM predictor but not support inference. For each method, we sample 1000 times based on the estimated joint and use samples to compute the marginals probabilities required by the GFM predictor, as described in Section~\ref{sec:support} (The CRF numbers are missing because there is no straight forward way of sampling from CRF). Comparing column ``L+E+G'' with column ``L+E'', we see that GFM alone always gives some improvement for BR and PCC, but is less effective for CBM. However, CBM clearly benefits from GFM when GFM is used in conjunction with support inference. Comparing all columns, we can conclude that support inference + GFM is the most effective prediction strategy, which consistently boosts the performance of all classifiers on all datasets.

\subsection{Other Related Work}
It is shown in \cite{ramaswamy2015design}  that the GFM algorithm can also work with non-probabilistic learners. For example, we could train a linear regression  with the target $\mathbbm{1}(y_l=1, |\mathbf{y}|=s\:|\:\mathbf{x})$, and the resulting  regression scores can be fed into GFM. 
Besides the GFM method  which works in the most general setting, there exist a few other F1 predictors specifically designed for BR \cite{chai2005expectation, jansche2007maximum, quevedo2012multilabel, nan2012optimizing}. There are also a few methods that take into account F1 metric during training.  \cite{parambath2014optimizing} reduces F metric maximization to cost-sensitive classification, and \cite{gasse2016f} studies F metric maximization with conditionally independent label subsets \cite{gasse2016f}.  \cite{pillai2017designing} provides an up-to-date review on different F metric maximization methods.

\begin{table}[h] \centering 

\begin{tabular}{c|c|c|c|c|c}
\hline
& MEDICAL & BIBTEX & OHSUMED &ENRON & WIPO\\
\hline 
SPEN &70.9 & 38.7 & 60.8 & 59.4 &64.0\\
\hline
CBM & 82.6 & 49.5 & 71.7 & 62.5 & 74.3\\
\hline
\end{tabular} 
  \caption{Test performance (F1) of Structured Prediction Energy Networks compared versus CBM} 
\label{tab:neural}
\end{table}
There are also several neural network based multi-label classification methods \cite{nam2014large, cisse2016adios, mnih2012conditional, belanger2016structured}. We ran the code associated with Structured Prediction Energy Networks (SPEN)~\cite{belanger2016structured} as it is without further adding the regularizations techniques discussed here, and the results are listed in Table~\ref{tab:neural} for reference. Granted that we did not spend as much time tuning the network parameters as we spent on the other methods, SPEN's performance looks less competitive, possibly due to over-fitting in high dimensional data with its high model capacity.

\section{Conclusion}
In this paper our main goal is to make effective F-measure optimal predictions on  multi-label text data. In doing so we show that most multi-label classification algorithms can be used if they produce a joint estimator $p(\mathbf{y} | \mathbf{x})$, to which we apply Support Inference ad GFM predictions, which work well together. In fact we show that on most datasets, classifiers both simple (BR) and complex (CRF) rely more on these two regularization ideas, rather than the their internal label dependency models (if any).

Working with text data, we also apply well established training regularization ideas: L1, L2, Early Stopping. We analyze the effects of these and notice significant reduction in model size (otherwise seriously big) and achieve state of the art performance on benchmark datasets. We make our code available at \url{https://github.com/cheng-li/pyramid}.



\bibliographystyle{ACM-Reference-Format}
\bibliography{local}


\begin{thebibliography}{00}


\ifx \showCODEN    \undefined \def \showCODEN     #1{\unskip}     \fi
\ifx \showDOI      \undefined \def \showDOI       #1{{\tt DOI:}\penalty0{#1}\ }
  \fi
\ifx \showISBNx    \undefined \def \showISBNx     #1{\unskip}     \fi
\ifx \showISBNxiii \undefined \def \showISBNxiii  #1{\unskip}     \fi
\ifx \showISSN     \undefined \def \showISSN      #1{\unskip}     \fi
\ifx \showLCCN     \undefined \def \showLCCN      #1{\unskip}     \fi
\ifx \shownote     \undefined \def \shownote      #1{#1}          \fi
\ifx \showarticletitle \undefined \def \showarticletitle #1{#1}   \fi
\ifx \showURL      \undefined \def \showURL       #1{#1}          \fi
\providecommand\bibfield[2]{#2}
\providecommand\bibinfo[2]{#2}
\providecommand\natexlab[1]{#1}
\providecommand\showeprint[2][]{arXiv:#2}

\bibitem[\protect\citeauthoryear{Belanger and McCallum}{Belanger and
  McCallum}{2016}]%
        {belanger2016structured}
\bibfield{author}{\bibinfo{person}{David Belanger} {and}
  \bibinfo{person}{Andrew McCallum}.} \bibinfo{year}{2016}\natexlab{}.
\newblock \showarticletitle{Structured prediction energy networks}. In
  \bibinfo{booktitle}{{\em Proceedings of the International Conference on
  Machine Learning}}.
\newblock


\bibitem[\protect\citeauthoryear{Chai}{Chai}{2005}]%
        {chai2005expectation}
\bibfield{author}{\bibinfo{person}{Kian Ming~Adam Chai}.}
  \bibinfo{year}{2005}\natexlab{}.
\newblock \showarticletitle{Expectation of F-measures: tractable exact
  computation and some empirical observations of its properties}. In
  \bibinfo{booktitle}{{\em Proceedings of the 28th annual international ACM
  SIGIR conference on Research and development in information retrieval}}. ACM,
  \bibinfo{pages}{593--594}.
\newblock


\bibitem[\protect\citeauthoryear{Ciss{\'e}, Al-Shedivat, and Bengio}{Ciss{\'e}
  et~al\mbox{.}}{2016}]%
        {cisse2016adios}
\bibfield{author}{\bibinfo{person}{Moustapha Ciss{\'e}},
  \bibinfo{person}{COM~Maruan Al-Shedivat}, {and} \bibinfo{person}{Samy
  Bengio}.} \bibinfo{year}{2016}\natexlab{}.
\newblock \showarticletitle{ADIOS: Architectures deep in output space}. In
  \bibinfo{booktitle}{{\em Proceedings of the 33rd International Conference on
  Machine Learning, ICML}}.
\newblock


\bibitem[\protect\citeauthoryear{Dembczy{\'n}ski, Waegeman, Cheng, and
  H{\"u}llermeier}{Dembczy{\'n}ski et~al\mbox{.}}{2012}]%
        {dembczynski2012label}
\bibfield{author}{\bibinfo{person}{Krzysztof Dembczy{\'n}ski},
  \bibinfo{person}{Willem Waegeman}, \bibinfo{person}{Weiwei Cheng}, {and}
  \bibinfo{person}{Eyke H{\"u}llermeier}.} \bibinfo{year}{2012}\natexlab{}.
\newblock \showarticletitle{On label dependence and loss minimization in
  multi-label classification}.
\newblock \bibinfo{journal}{{\em Machine Learning\/}} \bibinfo{volume}{88},
  \bibinfo{number}{1-2} (\bibinfo{year}{2012}), \bibinfo{pages}{5--45}.
\newblock


\bibitem[\protect\citeauthoryear{Friedman, Hastie, and Tibshirani}{Friedman
  et~al\mbox{.}}{2010}]%
        {friedman2010regularization}
\bibfield{author}{\bibinfo{person}{Jerome Friedman}, \bibinfo{person}{Trevor
  Hastie}, {and} \bibinfo{person}{Rob Tibshirani}.}
  \bibinfo{year}{2010}\natexlab{}.
\newblock \showarticletitle{Regularization paths for generalized linear models
  via coordinate descent}.
\newblock \bibinfo{journal}{{\em Journal of statistical software\/}}
  \bibinfo{volume}{33}, \bibinfo{number}{1} (\bibinfo{year}{2010}),
  \bibinfo{pages}{1}.
\newblock


\bibitem[\protect\citeauthoryear{Gasse and Aussem}{Gasse and Aussem}{2016}]%
        {gasse2016f}
\bibfield{author}{\bibinfo{person}{Maxime Gasse} {and} \bibinfo{person}{Alex
  Aussem}.} \bibinfo{year}{2016}\natexlab{}.
\newblock \showarticletitle{F-Measure Maximization in Multi-Label
  Classification with Conditionally Independent Label Subsets}. In
  \bibinfo{booktitle}{{\em Joint European Conference on Machine Learning and
  Knowledge Discovery in Databases}}. Springer, \bibinfo{pages}{619--631}.
\newblock


\bibitem[\protect\citeauthoryear{Ghamrawi and McCallum}{Ghamrawi and
  McCallum}{2005}]%
        {ghamrawi2005collective}
\bibfield{author}{\bibinfo{person}{Nadia Ghamrawi} {and}
  \bibinfo{person}{Andrew McCallum}.} \bibinfo{year}{2005}\natexlab{}.
\newblock \showarticletitle{Collective multi-label classification}. In
  \bibinfo{booktitle}{{\em Proceedings of the 14th ACM international conference
  on Information and knowledge management}}. ACM, \bibinfo{pages}{195--200}.
\newblock


\bibitem[\protect\citeauthoryear{Goodfellow, Bengio, and Courville}{Goodfellow
  et~al\mbox{.}}{2016}]%
        {goodfellow2016deep}
\bibfield{author}{\bibinfo{person}{Ian Goodfellow}, \bibinfo{person}{Yoshua
  Bengio}, {and} \bibinfo{person}{Aaron Courville}.}
  \bibinfo{year}{2016}\natexlab{}.
\newblock \bibinfo{booktitle}{{\em Deep learning}}.
\newblock \bibinfo{publisher}{MIT Press}.
\newblock


\bibitem[\protect\citeauthoryear{Jansche}{Jansche}{2007}]%
        {jansche2007maximum}
\bibfield{author}{\bibinfo{person}{Martin Jansche}.}
  \bibinfo{year}{2007}\natexlab{}.
\newblock \showarticletitle{A maximum expected utility framework for binary
  sequence labeling}. In \bibinfo{booktitle}{{\em Annual Meeting-Association
  For Computational Linguistics}}, Vol.~\bibinfo{volume}{45}.
  \bibinfo{pages}{736}.
\newblock


\bibitem[\protect\citeauthoryear{Kumar, Vembu, Menon, and Elkan}{Kumar
  et~al\mbox{.}}{2013}]%
        {kumar2013beam}
\bibfield{author}{\bibinfo{person}{Abhishek Kumar}, \bibinfo{person}{Shankar
  Vembu}, \bibinfo{person}{Aditya~Krishna Menon}, {and}
  \bibinfo{person}{Charles Elkan}.} \bibinfo{year}{2013}\natexlab{}.
\newblock \showarticletitle{Beam search algorithms for multilabel learning}.
\newblock \bibinfo{journal}{{\em Machine learning\/}} \bibinfo{volume}{92},
  \bibinfo{number}{1} (\bibinfo{year}{2013}), \bibinfo{pages}{65--89}.
\newblock


\bibitem[\protect\citeauthoryear{Lavergne, Capp{\'e}, and Yvon}{Lavergne
  et~al\mbox{.}}{2010}]%
        {lavergne2010practical}
\bibfield{author}{\bibinfo{person}{Thomas Lavergne}, \bibinfo{person}{Olivier
  Capp{\'e}}, {and} \bibinfo{person}{Fran{\c{c}}ois Yvon}.}
  \bibinfo{year}{2010}\natexlab{}.
\newblock \showarticletitle{Practical very large scale CRFs}. In
  \bibinfo{booktitle}{{\em Proceedings of the 48th Annual Meeting of the
  Association for Computational Linguistics}}. Association for Computational
  Linguistics, \bibinfo{pages}{504--513}.
\newblock


\bibitem[\protect\citeauthoryear{Li, Wang, Pavlu, and Aslam}{Li
  et~al\mbox{.}}{2016}]%
        {li2016conditional}
\bibfield{author}{\bibinfo{person}{Cheng Li}, \bibinfo{person}{Bingyu Wang},
  \bibinfo{person}{Virgil Pavlu}, {and} \bibinfo{person}{Javed~A. Aslam}.}
  \bibinfo{year}{2016}\natexlab{}.
\newblock \showarticletitle{Conditional Bernoulli Mixtures for Multi-label
  Classification}. In \bibinfo{booktitle}{{\em Proceedings of the 33rd
  International Conference on Machine Learning}}. \bibinfo{pages}{2482--2491}.
\newblock


\bibitem[\protect\citeauthoryear{Mnih, Larochelle, and Hinton}{Mnih
  et~al\mbox{.}}{2012}]%
        {mnih2012conditional}
\bibfield{author}{\bibinfo{person}{Volodymyr Mnih}, \bibinfo{person}{Hugo
  Larochelle}, {and} \bibinfo{person}{Geoffrey~E Hinton}.}
  \bibinfo{year}{2012}\natexlab{}.
\newblock \showarticletitle{Conditional restricted boltzmann machines for
  structured output prediction}.
\newblock \bibinfo{journal}{{\em arXiv preprint arXiv:1202.3748\/}}
  (\bibinfo{year}{2012}).
\newblock


\bibitem[\protect\citeauthoryear{Nam, Kim, Menc{\'\i}a, Gurevych, and
  F{\"u}rnkranz}{Nam et~al\mbox{.}}{2014}]%
        {nam2014large}
\bibfield{author}{\bibinfo{person}{Jinseok Nam}, \bibinfo{person}{Jungi Kim},
  \bibinfo{person}{Eneldo~Loza Menc{\'\i}a}, \bibinfo{person}{Iryna Gurevych},
  {and} \bibinfo{person}{Johannes F{\"u}rnkranz}.}
  \bibinfo{year}{2014}\natexlab{}.
\newblock \showarticletitle{Large-scale multi-label text
  classification—revisiting neural networks}. In \bibinfo{booktitle}{{\em
  Joint European Conference on Machine Learning and Knowledge Discovery in
  Databases}}. Springer, \bibinfo{pages}{437--452}.
\newblock


\bibitem[\protect\citeauthoryear{Nan, Chai, Lee, and Chieu}{Nan
  et~al\mbox{.}}{2012}]%
        {nan2012optimizing}
\bibfield{author}{\bibinfo{person}{Ye Nan}, \bibinfo{person}{Kian~Ming Chai},
  \bibinfo{person}{Wee~Sun Lee}, {and} \bibinfo{person}{Hai~Leong Chieu}.}
  \bibinfo{year}{2012}\natexlab{}.
\newblock \showarticletitle{Optimizing f-measure: A tale of two approaches}.
\newblock \bibinfo{journal}{{\em arXiv preprint arXiv:1206.4625\/}}
  (\bibinfo{year}{2012}).
\newblock


\bibitem[\protect\citeauthoryear{Parambath, Usunier, and Grandvalet}{Parambath
  et~al\mbox{.}}{2014}]%
        {parambath2014optimizing}
\bibfield{author}{\bibinfo{person}{Shameem~Puthiya Parambath},
  \bibinfo{person}{Nicolas Usunier}, {and} \bibinfo{person}{Yves Grandvalet}.}
  \bibinfo{year}{2014}\natexlab{}.
\newblock \showarticletitle{Optimizing F-measures by cost-sensitive
  classification}. In \bibinfo{booktitle}{{\em Advances in Neural Information
  Processing Systems}}. \bibinfo{pages}{2123--2131}.
\newblock


\bibitem[\protect\citeauthoryear{Pillai, Fumera, and Roli}{Pillai
  et~al\mbox{.}}{2017}]%
        {pillai2017designing}
\bibfield{author}{\bibinfo{person}{Ignazio Pillai}, \bibinfo{person}{Giorgio
  Fumera}, {and} \bibinfo{person}{Fabio Roli}.}
  \bibinfo{year}{2017}\natexlab{}.
\newblock \showarticletitle{Designing multi-label classifiers that maximize F
  measures: State of the art}.
\newblock \bibinfo{journal}{{\em Pattern Recognition\/}}  \bibinfo{volume}{61}
  (\bibinfo{year}{2017}), \bibinfo{pages}{394--404}.
\newblock


\bibitem[\protect\citeauthoryear{Quevedo, Luaces, and Bahamonde}{Quevedo
  et~al\mbox{.}}{2012}]%
        {quevedo2012multilabel}
\bibfield{author}{\bibinfo{person}{Jos{\'e}~Ram{\'o}N Quevedo},
  \bibinfo{person}{Oscar Luaces}, {and} \bibinfo{person}{Antonio Bahamonde}.}
  \bibinfo{year}{2012}\natexlab{}.
\newblock \showarticletitle{Multilabel classifiers with a probabilistic
  thresholding strategy}.
\newblock \bibinfo{journal}{{\em Pattern Recognition\/}} \bibinfo{volume}{45},
  \bibinfo{number}{2} (\bibinfo{year}{2012}), \bibinfo{pages}{876--883}.
\newblock


\bibitem[\protect\citeauthoryear{Ramaswamy}{Ramaswamy}{2015}]%
        {ramaswamy2015design}
\bibfield{author}{\bibinfo{person}{Harish~Guruprasad Ramaswamy}.}
  \bibinfo{year}{2015}\natexlab{}.
\newblock {\em \bibinfo{title}{Design and Analysis of Consistent Algorithms for
  Multiclass Learning Problems}}.
\newblock \bibinfo{thesistype}{Ph.D. Dissertation}. \bibinfo{school}{Indian
  Institute of Science Bangalore}.
\newblock


\bibitem[\protect\citeauthoryear{Read, Pfahringer, Holmes, and Frank}{Read
  et~al\mbox{.}}{2011}]%
        {read2011classifier}
\bibfield{author}{\bibinfo{person}{Jesse Read}, \bibinfo{person}{Bernhard
  Pfahringer}, \bibinfo{person}{Geoff Holmes}, {and} \bibinfo{person}{Eibe
  Frank}.} \bibinfo{year}{2011}\natexlab{}.
\newblock \showarticletitle{Classifier chains for multi-label classification}.
\newblock \bibinfo{journal}{{\em Machine learning\/}} \bibinfo{volume}{85},
  \bibinfo{number}{3} (\bibinfo{year}{2011}), \bibinfo{pages}{333--359}.
\newblock


\bibitem[\protect\citeauthoryear{Tsoumakas and Katakis}{Tsoumakas and
  Katakis}{2007}]%
        {tsoumakas2006multi}
\bibfield{author}{\bibinfo{person}{Grigorios Tsoumakas} {and}
  \bibinfo{person}{Ioannis Katakis}.} \bibinfo{year}{2007}\natexlab{}.
\newblock \showarticletitle{Multi-label classification: An overview}.
\newblock \bibinfo{journal}{{\em Int J Data Warehousing and Mining\/}}
  \bibinfo{volume}{2007} (\bibinfo{year}{2007}), \bibinfo{pages}{1--13}.
\newblock


\bibitem[\protect\citeauthoryear{Waegeman, Dembczy{\'n}ki, Jachnik, Cheng, and
  H{\"u}llermeier}{Waegeman et~al\mbox{.}}{2014}]%
        {waegeman2014bayes}
\bibfield{author}{\bibinfo{person}{Willem Waegeman}, \bibinfo{person}{Krzysztof
  Dembczy{\'n}ki}, \bibinfo{person}{Arkadiusz Jachnik}, \bibinfo{person}{Weiwei
  Cheng}, {and} \bibinfo{person}{Eyke H{\"u}llermeier}.}
  \bibinfo{year}{2014}\natexlab{}.
\newblock \showarticletitle{On the bayes-optimality of F-measure maximizers}.
\newblock \bibinfo{journal}{{\em The Journal of Machine Learning Research\/}}
  \bibinfo{volume}{15}, \bibinfo{number}{1} (\bibinfo{year}{2014}),
  \bibinfo{pages}{3333--3388}.
\newblock


\bibitem[\protect\citeauthoryear{Yuan, Ho, and Lin}{Yuan et~al\mbox{.}}{2012}]%
        {yuan2012improved}
\bibfield{author}{\bibinfo{person}{Guo-Xun Yuan}, \bibinfo{person}{Chia-Hua
  Ho}, {and} \bibinfo{person}{Chih-Jen Lin}.} \bibinfo{year}{2012}\natexlab{}.
\newblock \showarticletitle{An improved glmnet for l1-regularized logistic
  regression}.
\newblock \bibinfo{journal}{{\em Journal of Machine Learning Research\/}}
  \bibinfo{volume}{13}, \bibinfo{number}{Jun} (\bibinfo{year}{2012}),
  \bibinfo{pages}{1999--2030}.
\newblock


\end{thebibliography}

\end{document}